\PassOptionsToPackage{expansion=false}{microtype}
\documentclass[sigconf,nonacm]{acmart} 
\usepackage[ruled,vlined]{algorithm2e}

\hypersetup{draft} 

\AtBeginDocument{%
  }

\setcopyright{none}
\renewcommand\footnotetextcopyrightpermission[1]{}

\begin{document}

\title{SLA-Constrained Carbon-Aware Routing in Geo-Distributed Serverless Clouds}


% \correspondingauthor
% \affiliation{%

% \affiliation{%

% \affiliation{%

% \affiliation{%

% \affiliation{%

% \affiliation{%

% \affiliation{%

\author{Anmol Chaudhary}
% \correspondingauthor
\affiliation{
  \institution{National Institute of Advanced Manufacturing Technology}
  \city{Ranchi}
  \country{India}
}
\email{anmolc563@gmail.com}

\author{Rahul Mishra}
\affiliation{
   \institution{National Institute of Advanced Manufacturing Technology}
  \city{Ranchi}
  \country{India}
}
\email{rmishra@niamt.ac.in}

\begin{abstract}
Modern cloud deployments distribute applications across multiple geographic regions, yet standard routing mechanisms prioritize latency while ignoring the fluctuating carbon intensity of local power grids. Latency-driven routing incurs avoidable carbon emissions, particularly when cleaner regions are within acceptable latency bounds. The proposed model formulates the carbon-aware serverless routing problem as a constrained optimization over geo-distributed cloud regions and introduces an SLA-constrained carbon-aware routing policy that achieves optimal carbon reduction within the SLA-feasible region, evaluated using real carbon intensity measurements across 5 primary AWS deployments. Experimental results show that the proposed policy achieves up to 46.8\% carbon reduction while maintaining zero SLA violations across all evaluated thresholds. The system reduces carbon by an average of 27.4\% under mixed workloads, and the routing overhead is very low (less than 0.02\% of total request latency). A scalability study across 12 AWS regions spanning 6 continents demonstrates that average carbon savings increase from 27.4\% to 47.5\% as routing flexibility expands under mixed workloads. The proposed work contributes to SDG 13 (Climate Action) and SDG 7 (Affordable and Clean Energy) by enabling low-carbon routing decisions. These results indicate that cloud systems can achieve significant carbon savings without compromising user experience.
\end{abstract}


\maketitle

\section{Introduction}
Cloud computing platforms are progressively relying on geographically distributed data centers and edge servers to reduce latency, improve performance, and meet the demands of global, real-time applications. Currently, the request routing decisions are primarily determined based on the latency requirements, which generally maps to the region closest to the users. While the current approach provides the optimal end-to-end latency, it entirely ignores the carbon footprint of the selected region, i.e., the carbon intensity (gCO$_2$/kWh) of the electricity grid powering that region's data center at the time of execution. As carbon intensity varies significantly across locations and time, latency-only routing can result in avoidable carbon emissions even when cleaner alternatives exist within acceptable latency bounds.
Recently, researchers have explored the concept of green computing, which aims to reduce the carbon footprint of data centers \cite{carbonexplorer2023, beloglazov2012}. While prior work has begun incorporating carbon awareness into serverless scheduling \cite{casa2024,greencourier2023} and spatial shifting across VM and microservice infrastructure \cite{casper2023,aceso2026,maji2023}, none of these systems simultaneously enforce per-request SLA constraints as a hard feasibility boundary, operate on managed serverless platforms without infrastructure modification, and demonstrate deployment on real cloud infrastructure.

\subsection{Motivation \& Contributions}
The rapid growth of cloud-native and serverless computing has created a significant impact on the carbon footprint of distributed computing. Serverless computing has emerged as a dominant paradigm. Scalability, cost-effectiveness, and ease of development are major features of serverless computing. Although technological and financial aspects are well understood, environmental implications are less explored \cite{awwad2025estimating}. The increased need for computing and its infrastructure may lead to an increased carbon footprint \cite{akour2025reducing,asadov2025carbon}.

Moreover, the carbon footprint of the electrical supply system varies significantly over different regions, depending on the diversity of the mix, penetration levels of renewable energy sources, and time-of-use patterns. The variability in grid carbon intensity creates an opportunity to reduce the carbon footprint through intelligent request routing to regions with a smaller carbon footprint, while the latency requirements are satisfied properly to avoid the deterioration of the quality of service.

The carbon footprint of cloud applications is not limited to compute; recent advancements have even begun exploring dynamic frequency scaling on network switches coupled with carbon-aware routing to mitigate the emissions of wide-area data transit itself \cite{10.1145/3768984}.

%% NEW PARAGRAPH — addresses R1's "contribution vs related work unclear" %%

While carbon-aware spatial shifting has been explored for persistent virtual machine and microservice infrastructure \cite{casper2023,aceso2026}, these systems are designed around long-running, migratable workloads and treat latency as one term in a jointly weighted objective rather than as a hard service guarantee. Serverless functions, by contrast, are short-lived, stateless, and typically governed by strict per-request latency SLAs that cannot be relaxed without directly degrading user-facing quality of service. The underlying distinction motivates a routing formulation in which the SLA is enforced as a hard feasibility constraint, and carbon is minimized only within the resulting feasible region, rather than traded off against latency through weighting.

In the proposed work, the gap is addressed by proposing an SLA-constrained carbon routing framework for serverless applications. The core design principle is that, instead of routing to the closest region or the region with the lowest carbon footprint, the system first routes to the region that meets the latency requirement, then to the region with the lowest carbon intensity among them. The approach ensures that carbon reduction is achieved only when it does not compromise user experience \cite{sharma2023challenges}.

The proposed model is implemented and evaluated across 5 primary AWS regions, and a scalability study was performed up to 12 regions across 6 continents using real-world carbon intensity data obtained from Electricity Maps and empirically measured network latencies.

The 5 primary regions are selected to cover a wide range of carbon intensities and physical distances from the request's source, and the additional 7 regions used in the scalability study are selected to provide coverage over all 6 populated continents, ensuring the evaluation is not limited to a geographically or climatically narrow subset.

To systematically analyze the trade-offs, five routing policies are compared:

\begin{enumerate}
\item A baseline latency-driven policy selecting the nearest region.
\item A carbon-optimal policy that ignores latency constraints.
\item The proposed SLA-aware carbon routing policy.
\item A cold start-aware variant incorporating first-invocation penalties.

%% MODIFIED — softened claim, addresses R1's CASPER misrepresentation concern %% 

\item A weighted-scoring ablation baseline inspired by 
spatial-shifting approaches such as CASPER \cite{casper2023}, used to isolate the effect of 
hard versus soft SLA enforcement.

% \item \textbf{A weighted-scoring baseline structurally inspired by spatial-shifting approaches such as CASPER \cite{casper2023}, used here as a controlled ablation to isolate the effect of hard SLA enforcement versus soft latency weighting.}
\end{enumerate}

The proposed work contributes to Sustainable Development Goal (SDG) 13: Climate Action by reducing the carbon footprint of cloud applications through smart routing decisions. Furthermore, it supports SDG 7: Affordable and Clean Energy by promoting the utilization of low-carbon regions.

The contributions of the proposed work are as follows:

\begin{enumerate}
\item We formulate serverless request routing as an SLA constrained carbon minimization problem. We propose an SLA-constrained carbon-aware routing policy that achieves up to 46.8\% carbon reduction with zero SLA violations.

\item The proposed work demonstrates that managed serverless platforms can achieve significant carbon savings without requiring modifications to application logic or cloud infrastructure. It enables real-world, deployment-transparent sustainability.

%% NEW CONTRIBUTION — addresses R1's "contribution unclear" directly in the list, not just in prose above %%
\item Unlike prior spatial-shifting systems targeting VM or microservice infrastructure, the proposed work enforces per-request SLA feasibility as a hard constraint rather than as a weighted objective, as empirically demonstrated through a controlled ablation study.

\item We introduce an experimentally derived cold-start penalty model ($\alpha$ = 0.8) based on real AWS Lambda first-invocation measurements and show that the proposed policy remains robust under worst-case cold-start conditions.

%% MODIFIED — softened "CASPER-inspired baseline" framing to match the ablation reframe %% 

\item Using a soft-weighted baseline inspired by spatial-shifting methods like CASPER \cite{casper2023}, we show that shifting to hard SLA filtering boosts carbon savings by up to 12.8\% while completely wiping out the high SLA violations caused by soft optimization. 
\end{enumerate}

\section{Related Work}

Carbon-aware computing techniques for cloud infrastructure can be broadly organized along two axes: when a workload is executed (temporal shifting) and where it is executed (spatial shifting), with a more recent line of work addressing the distinct challenges posed by serverless execution. Temporal approaches, including GREEN \cite{green2025}, CarbonScaler \cite{carbonscaler2024}, and the limitations analysis by T. Sukprasert et al. \cite{sukprasert2024limitations} reduce carbon emissions by delaying batch workloads until low-carbon periods. However, such delay-based strategies are fundamentally incompatible with latency-sensitive applications governed by strict SLAs.

Where temporal approaches are unsuitable, spatial shifting offers an alternative for latency-sensitive workloads by relocating execution rather than delaying it. Spatial shifting routes workloads to regions with lower carbon intensity, and systems such as CASPER \cite{casper2023}, Aceso \cite{aceso2026}, and Caribou \cite{gsteiger2024caribou} model it as a multi-objective optimization problem balancing carbon emissions and performance. CASPER further incorporates service-level objectives (SLOs) and maximum-latency thresholds as constraints within its optimization. However, recent work on multi-cloud carbon-aware routing operates at the level of virtual machines or microservices and does not address per-request routing for serverless workloads under strict SLA constraints \cite{maji2023}.

%% NEW TRANSITION SENTENCE — connects spatial-for-VMs paragraph to geo-distributed scheduling paragraph %%
Moreover, for carbon-aware systems, geo-distributed scheduling has generally addressed similar placement decisions for both performance and cost, but without considering carbon-emissions objectives. Geo-distributed scheduling has been examined for performance, cost, and resource optimization, including approaches based on Lyapunov optimization \cite{zhou2013} and reinforcement learning. However, none of the mentioned work focuses on serverless computing and assumes that carbon emissions are a competing objective of latency in their models. 

Serverless computing introduces new challenges due to its event-driven nature, i.e. cold start. Existing work primarily focuses on mitigating cold starts and improving resource allocation. Systems such as StepConf \cite{stepconf2022}, CASA \cite{casa2024}, and EcoLife \cite{eclife2024} focus on performance and carbon-aware scheduling, while GreenCourier \cite{greencourier2023} and GreenWhisk \cite{greenwhisk2024} explore carbon-aware execution. However, these systems are largely intra-regional and often require modifications to underlying platforms, limiting their applicability to managed services such as AWS Lambda. Also, the recent work explores multi-objective optimization frameworks incorporating carbon, latency, and cost, often using reinforcement learning or microservice placement strategies such as Aceso \cite{aceso2026}. However, none of these support per-request routing in real-world deployments.

Thus, from the above, the significant key gaps can be identified: (i) reliance of existing systems on temporal shifting or relaxed latency constraints; (ii) geo-distributed scheduling is tailored to persistent infrastructure; and (iii) lack of inter-region, per-request routing for serverless workloads. The proposed work directly addresses these gaps by enabling SLA-aware, per-request, carbon-efficient routing across regions for serverless workloads. Unlike prior approaches, we enforce the SLA as a hard feasibility constraint rather than incorporating it as a weighted term in the optimization objective. Furthermore, the proposed approach requires no modifications to the underlying infrastructure and is evaluated on real-world latency and carbon-intensity traces.

Table \ref{tab:related} compares existing work with the proposed system across six parameters: carbon awareness, SLA conformance, deployment realism, serverless support, per-request routing, and infrastructure independence. Among these, CASPER \cite{casper2023} and Aceso \cite{aceso2026} are the prior systems that are simultaneously carbon- and SLA-aware. However, neither targets the serverless execution model nor supports per-request routing, and both require modifications to the underlying scheduling infrastructure.

\begin{table}[htbp]
\centering
\caption{Comparison of Existing Carbon-Aware Systems with the 
Proposed Framework across Key Architectural Dimensions}
\label{tab:related}
\resizebox{\columnwidth}{!}{%
\small
\begin{tabular}{|l|c|c|c|c|c|c|}
\hline
\textbf{Work} & 
\textbf{\shortstack{Carbon\\Aware}} & 
\textbf{\shortstack{SLA\\Aware}} & 
\textbf{\shortstack{Real\\Deploy}} & 
\textbf{\shortstack{Server\\less}} & 
\textbf{\shortstack{Per-Req\\Routing}} & 
\textbf{\shortstack{No-Infra\\Modification}} \\
\hline
CASPER \cite{casper2023} & Yes & Yes & No & No & No & No \\ \hline
Maji et al. \cite{maji2023} & Yes & Yes & Yes & No & Yes & Partial \\ \hline
GREEN \cite{green2025} & Yes & No & No & No & No & No \\ \hline
Aceso \cite{aceso2026} & Yes & Yes & No & No & No & No \\ \hline
GreenCourier \cite{greencourier2023} & Yes & No & Ltd & Yes & Partial & No \\ \hline
CASA \cite{casa2024} & Yes & Yes & No & Yes & Partial & No \\ \hline
\textbf{Proposed} & \textbf{Yes} & \textbf{Yes} & \textbf{Yes} & \textbf{Yes} & \textbf{Yes} & \textbf{Yes} \\ 
\hline
\end{tabular}%
}
\end{table}

\section{System Design}
The proposed framework enables carbon-aware serverless request routing by combining latency-aware region selection with real-time carbon intensity data. This section presents the system architecture, routing workflow, optimization model, and implementation details.
\subsection{System Overview}
The proposed work introduces a carbon-aware routing system for serverless applications across different cloud regions. A request from the client is received at the API Gateway, which forwards it to a carbon-aware router, implemented as an AWS Lambda function. The router does not execute the actual computation. Instead, it decides the best region for the execution of the function based on the latency requirements and carbon intensity. If no region is found that meets the SLA (Service Level Agreement) requirements, the request is routed to the nearest region as a fallback.
The router is based on two key inputs:
(i) a precomputed latency profile that includes the round-trip time (RTT) from the client to each cloud region, and
(ii) a carbon intensity dataset with an hourly update frequency.
Based on the above inputs, the router determines the target region and routes the request to an identical Lambda function running in that region.

\begin{figure}[htbp]
\centering
\includegraphics[width=8cm,height=5cm]{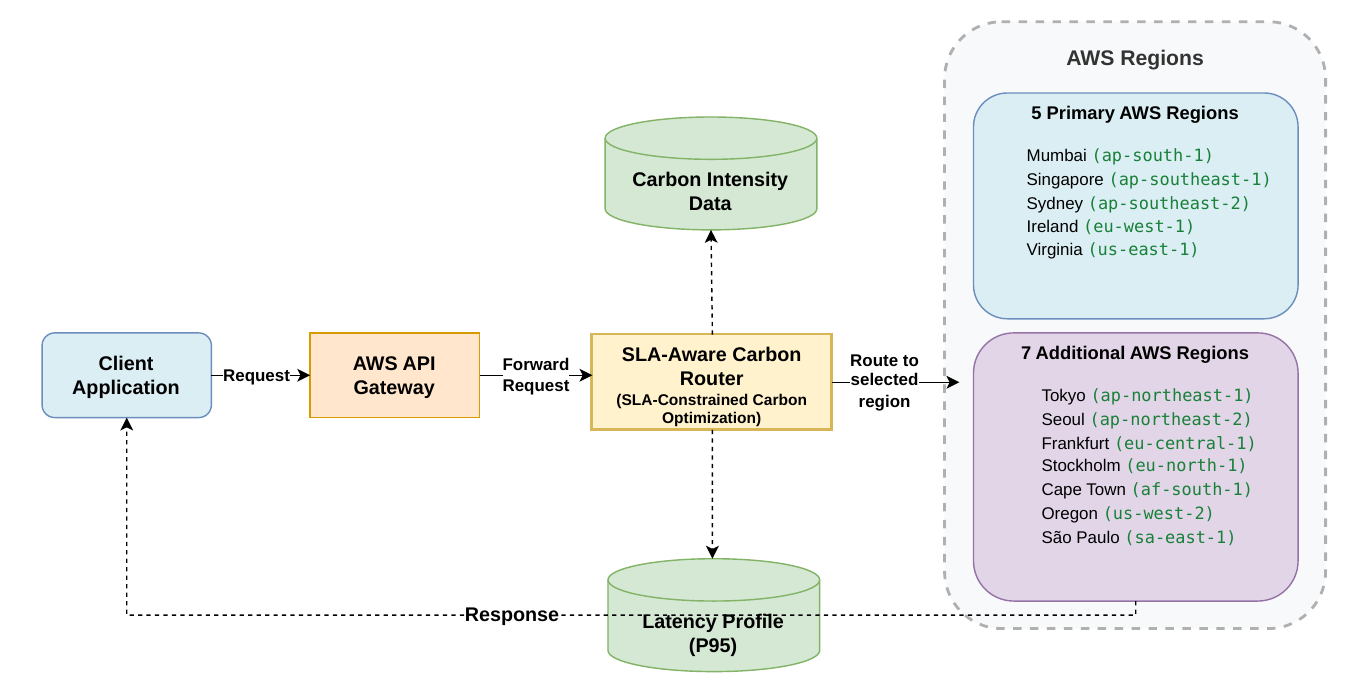}
\caption{Architecture of the Proposed SLA-Aware Carbon Routing Framework}
\label{fig:architecture}
\end{figure}

The end-to-end architecture of the proposed system is outlined in Figure \ref{fig:architecture}. Incoming client requests are routed through AWS API Gateway and then forwarded to the Carbon Aware Router Lambda function. The routing decision is made based on the carbon intensity and the P95 latency profile in less than 0.036ms. The request is forwarded to one of the 5 geographically distributed AWS Lambda regions: Mumbai, Singapore, Sydney, Ireland, or Virginia, with a scalability study extended to 12 regions across 6 continents. The response is returned directly to the client, bypassing the router entirely.

\subsection{Routing Decision}
For each incoming request, the router extracts the SLA requirement and performs a constant-time lookup on both the latency and carbon datasets. Let $R$ denote the set of all regions. The router computes a feasible subset:

\begin{equation*}
R_{SLA} = \{ r \in R \mid L(r) \leq SLA \}
\end{equation*}

Further, the selected region $r^*$ is computed as

$$
 r^* = \arg\min_{r \in R_{SLA}} C(r,t)
$$

The routing decision can be formally expressed as:

\[
r^{*} \equiv
\begin{cases}
\arg\min_{r \in R_{SLA}} C(r,t)  & \text{if } R_{SLA} \neq \emptyset, \\
\arg\min_{r \in R} L(r)  & \text{otherwise.}
\end{cases}
\]

\begin{figure}[htbp]
\centering
\includegraphics[width=\linewidth]{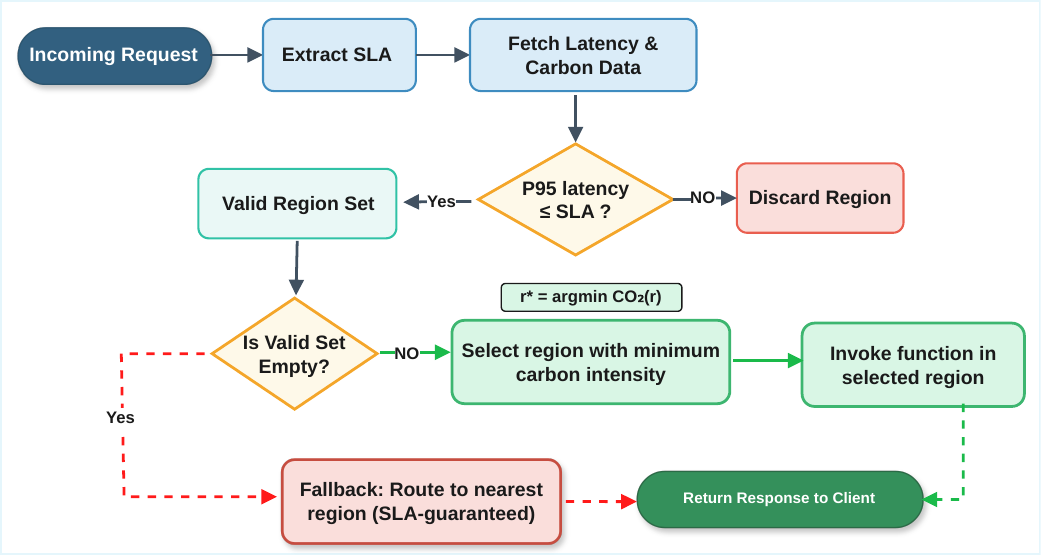}
\caption{SLA-Aware Carbon Routing Decision Workflow}
\label{fig:decision_maker}
\end{figure}

Figure \ref{fig:decision_maker} describes the workflow of routing decisions of the proposed SLA-aware carbon routing policy. First, the router evaluates all candidate regions for each incoming request and creates a feasible set of regions with an SLA latency constraint based on P95 latency measurements. If the feasible set is non-empty, the router selects the region with the lowest carbon intensity at the current time. In case where there is no feasible set, the fallback mechanism is employed for choosing the closest region to ensure the SLA is fulfilled. The selected region will perform the request and reply back to the client. The decision process is formally described in Algorithm \ref{alg:sla_routing}.

\begin{algorithm}
\caption{SLA-Aware Carbon Routing}
\label{alg:sla_routing}
\KwIn{Regions $R$, latency $L(r)$, carbon intensity $C(r,t)$, SLA}
\KwOut{Selected region $r^*$}

$R_{SLA} \leftarrow \{ r \in R \mid L(r) \leq SLA \}$\;

\If{$R_{SLA} = \emptyset$}{
    \Return $\arg\min_{r \in R} L(r)$ \tcp*{Fallback to nearest region}
}

\Return $\arg\min_{r \in R_{SLA}} C(r,t)$\;

\end{algorithm}

The routing decision involves a single pass over the set of regions, resulting in a time complexity of $O(|R|)$. As $|R|$ is small and bounded in our scenario, the decision process runs in constant time with negligible overhead.

\subsection{Problem Formulation}
The routing decision is treated as a constrained optimization problem on a set of regions $R$. For each incoming request, the goal is to minimize the carbon intensity under a constraint on the latency requirement.

Let:
\begin{itemize}
\item $L(r)$ denote the P95 latency to region $r$
\item $C(r,t)$ denote the carbon intensity of region $r$ at time $t$
\item $SLA$ denote the latency constraint for the request
\end{itemize}

The routing decision is formulated as:

\begin{equation}
\min_{r \in R} \; C(r, t)
\end{equation}

subject to:

\begin{equation}
L(r) \leq SLA
\end{equation}

If no region satisfies the SLA constraint, the system falls back to selecting the region with minimum latency:

\begin{equation}
r^* = \arg\min_{r \in R} L(r)
\end{equation}

The formulation ensures that carbon optimization is performed only within the feasible latency region, preserving user experience.

By treating the SLA as a hard feasibility boundary rather than a weighted objective, the formulation guarantees zero SLA violations by construction whenever a feasible region exists.

\subsection{Implementation Details}
The Carbon Aware Router is implemented as a lightweight AWS Lambda function running in the Mumbai region. The carbon intensity data is locally cached from a precomputed dataset from Electricity Maps (January 2025), eliminating runtime API dependency and enabling low-latency routing. Real-time carbon feed integration is identified as a direction for future work. 

% The latency figures are based on empirical measurements and stored in a lookup table.

The routing decision is implemented using a lightweight, in-memory evaluation over precomputed latency and carbon profiles, enabling efficient per-request carbon optimization with negligible routing overhead.

As shown in section \ref{result_section}, the latency of decision remains below 0.036 ms (P99) and represents less than 0.02\% of the end-to-end latency. The same workload function is used across all regions to ensure consistent execution and fair comparison. The deployed function performs a fixed-cost JSON echo to verify the measured differences across policies reflect routing decisions rather than workload variance.

To simulate the realistic traffic, a mixed workload across 5 SLA categories is evaluated: real-time (300 ms), interactive (500 ms), moderate (800 ms), relaxed (1000 ms), and batch (3000 ms), each routed independently based on latency constraints and carbon intensity.

\section{Experimental Setup}
The evaluation of the proposed routing framework utilizes geographically distributed AWS regions, real-world carbon intensity data, and p95 latency profiles. The section outlines the deployment infrastructure, latency measurement methodology, workload configuration, and the various routing policies implemented to evaluate the effectiveness of the proposed approach.

\subsection{Deployment Infrastructure and Carbon Intensity Data}

Experiments have been performed on Amazon Web Services (AWS) leveraging 5 geographic regions: ap-south-1 (Mumbai), ap-southeast-1 (Singapore), ap-southeast-2 (Sydney), eu-west-1 (Ireland), and us-east-1 (Virginia). These 5 regions were selected to cover a wide range of network distances and grid carbon intensities relative to the request origin. In all of these regions, the same serverless function has been instantiated using AWS Lambda with Python 3.13 and 128 MB memory. The function performs a lightweight task, ensuring consistent execution across regions. For the scalability study, the same function was additionally deployed across seven more regions: eu-north-1 (Stockholm), ap-northeast-2 (Seoul), af-south-1 (Cape Town), eu-central-1 (Frankfurt), sa-east-1 (São Paulo), ap-northeast-1 (Tokyo), and us-west-2 (Oregon), spanning 6 continents in total. These additional regions were chosen to extend continental coverage rather than to further refine carbon-intensity granularity within any single region already represented.

\begin{table}[htbp]
\centering
\caption{Experimental Setup}
\label{tab:setup}
\resizebox{\columnwidth}{!}{%
\small
\begin{tabular}{|l|l|}
\hline
\textbf{Parameter} & \textbf{Value} \\ \hline

Cloud Provider & Amazon Web Services (AWS) \\ \hline

\textbf{5 Primary Regions} & \shortstack[l]{
    ap-south-1 (Mumbai), ap-southeast-1 (Singapore),\\
    ap-southeast-2 (Sydney), eu-west-1 (Ireland),\\
    us-east-1 (Virginia)
} \\ \hline

\textbf{7 Additional Regions} & \shortstack[l]{
    ap-northeast-1 (Tokyo), ap-northeast-2 (Seoul),\\
    eu-central-1 (Frankfurt), eu-north-1 (Stockholm),\\
    af-south-1 (Cape Town), us-west-2 (Oregon),\\
    sa-east-1 (S\~ao Paulo)
} \\ \hline

Lambda Runtime & Python 3.13 \\ \hline
Lambda Memory & 128 MB \\ \hline
Carbon Data Source & Electricity Maps Dataset \\ \hline
Granularity & Hourly \\ \hline
Dataset Period & January 2025 (744 hours) \\ \hline
Simulation Runs & 30 per SLA \\ \hline
Requests per Run & 1000 (per-SLA), 5000 (mixed) \\ \hline
Statistical Test & Mann-Whitney U ($p < 0.05$) \\ \hline
Scalability Study & 12 regions, 6 continents (5$\rightarrow$12 incremental) \\ \hline

\end{tabular}%
}
\end{table}

Table \ref{tab:setup} highlights the experimental setup, including the cloud architecture, dataset properties, simulation settings, and Scalability Study.
The carbon intensity values for each region have been obtained from the Electricity Maps Dataset. The dataset comprises hourly carbon intensity measurements (gCO$_2$/kWh) for January 2025, yielding a total of 744 data points for each region.

\subsection{Latency Measurement}

For each region, the latency between the client and server was calculated by directly invoking the AWS Lambda function using the boto3 library. Instead of the mean value, the P95 value of the round-trip-time distribution is used throughout the study for SLA filtering, as it captures tail latency and provides a more conservative basis for ensuring SLA compliance. For each region, 20 invocations were made, and their RTTs were recorded after excluding the cold-start effect. The measured P95 round-trip latencies for the 5 primary AWS regions were: Mumbai (ap-south-1) 196 ms, Singapore (ap-southeast-1) 367 ms, Sydney (ap-southeast-2) 634 ms, Ireland (eu-west-1) 979 ms, and Virginia (us-east-1) 1491 ms. These values reflect significant geographical variation and form the basis for SLA filtering in the routing decision.

For the 7 additional regions, P95 latency values were rounded to the nearest 10 ms, as these regions were included solely for routing-pool scalability analysis rather than fine-grained SLA-threshold sensitivity evaluation. The corresponding P95 latencies were: Tokyo (ap-northeast-1) 600 ms, Seoul (ap-northeast-2) 620 ms, Frankfurt (eu-central-1) 720 ms, Stockholm (eu-north-1) 780 ms, Cape Town (af-south-1) 950 ms, Oregon (us-west-2) 1050 ms, and São Paulo (sa-east-1) 1150 ms. 

Cold-start exclusion was performed by discarding the first invocation of each measurement session and computing the P95 RTT over the remaining 19 invocations. The discarded first-invocation values were retained separately as cold-start penalty measurements for Policy C+, rather than being excluded from the dataset entirely. The measured first-invocation latencies were: Mumbai 423ms, Singapore 391ms, Sydney 952ms, Ireland 1108ms, and Virginia 1626ms. Similarly, first-invocation latencies for the seven scalability regions were measured using the same methodology and, where applicable, incorporated into the cold-start penalty model. Across all five regions and 20 invocations per region, cold start overhead was observed in 78–83\% of first-invocation attempts following a period of inactivity, yielding a mean cold start occurrence rate of 0.80. The mean cold-start occurrence rate of 0.80 is used as the conservatism parameter $\alpha = 0.8$ in Policy C+, ensuring that effective latency is never underestimated during region transitions.

\subsection{Workload and Simulation}
Routing performance is evaluated using simulated request workloads derived from the collected carbon and latency datasets. For each SLA threshold (300 ms, 500 ms, 800 ms, 1000 ms, and 3000 ms), 1000 requests were generated per run, and each experiment was repeated 30 times (n = 30,000 routing decisions per SLA threshold per policy) to ensure statistical reliability. Additionally, a mixed workload scenario consisting of 5000 requests per run was evaluated over 30 repetitions (n = 150,000 routing decisions), with requests distributed across SLA categories to reflect realistic traffic patterns.

\subsection{Routing Policies}

To assess the performance of the carbon-aware routing, five different routing policies are evaluated as comparators against the proposed approach:

\begin{enumerate}
    \item \textbf{Policy A (Latency-Optimal Baseline):} In Policy A, the requests are routed to the region that is nearest to the source (Mumbai in our case). Here, requests are minimizing latency but ignoring variations in carbon intensity across regions.

    \item \textbf{Policy B (Carbon-Optimal):} In Policy B, the requests are routed to the region with the lowest carbon intensity. The requests are directed to the region that minimizes the carbon intensity, irrespective of the latency constraint. It achieves maximum carbon savings but may lead to SLA violations.

    \item \textbf{Policy C (SLA-Aware Carbon Routing, Proposed):} Requests are routed to the carbon-minimizing region only within the SLA-feasible subset, in two phases: (1) \textit{SLA Feasibility Filtering:} candidate regions with P95 latency exceeding the per-request SLA threshold are eliminated, restricting routing to the latency-feasible set; (2) \textit{Carbon-Optimal Selection:} among feasible regions, the one with minimum instantaneous carbon intensity $C(r, t)$ is selected, ensuring carbon reduction without violating service-level guarantees.

    \item \textbf{Policy C+ (Cold Start-Aware Carbon Routing):} Policy C+ builds on Policy C by incorporating cold start penalties derived from measured first-invocation latencies. The effective latency is calculated as:
    \begin{equation}
    L_{\text{eff}}(r) = L_{\text{measured}}(r) + \alpha \cdot T_{\text{cold}}(r)
    \end{equation}
    where $T_{\text{cold}}(r)$ is the empirically measured first-invocation latency for region $r$, and $\alpha \in [0,1]$ is a conservatism parameter set to $0.8$, derived from the observed cold start occurrence rate of 78--83\% across first-invocation measurements in all 5 AWS regions. The conservative fixed value prevents the system from underestimating latency during region changes. It also prevents temporary increases in latency from causing SLA violations.

    \item \textbf{Policy D (Soft Weighted-Scoring Ablation):} Policy D implements a weighted scoring method based on carbon and latency. It is inspired by spatial-shifting framework such as CASPER \cite{casper2023}, but intentionally avoids strict SLA enforcement to isolate the effect of constraint enforcement against soft weighting, serving as a controlled ablation rather than a baseline comparison. The routing score is computed as:
    \begin{equation}
    \text{Score}(r) = w \cdot C_{\text{norm}}(r) + (1 - w) \cdot L_{\text{norm}}(r)
    \end{equation}
    where $w = 0.5$, and $C_{\text{norm}}(r)$ and $L_{\text{norm}}(r)$ are normalized carbon and latency values. The region with the minimum score is selected.
\end{enumerate}

The primary evaluation metric is carbon intensity (gCO$_2$/kWh) associated with each routed request. Secondary metrics are the SLA violation rate and routing overhead. Statistical analysis is performed using the Mann-Whitney U test ($p < 0.05$), which is applicable to non-normally distributed data (latency and carbon footprint). Results are reported as mean values over 30 independent runs per condition.

\section{Results and Evaluation} \label{result_section}
To experimentally evaluate the proposed SLA-aware carbon routing framework, real-world carbon intensity data and p95 latency profiles are used. The analysis covers carbon intensity, carbon savings across different SLA thresholds, routing behavior, statistical validation, mixed workload performance, scalability, and routing overhead.

\subsection{Carbon Intensity Variation}

\begin{figure*}[t]
\centering
\includegraphics[height=8.5cm,width=1.0\textwidth]{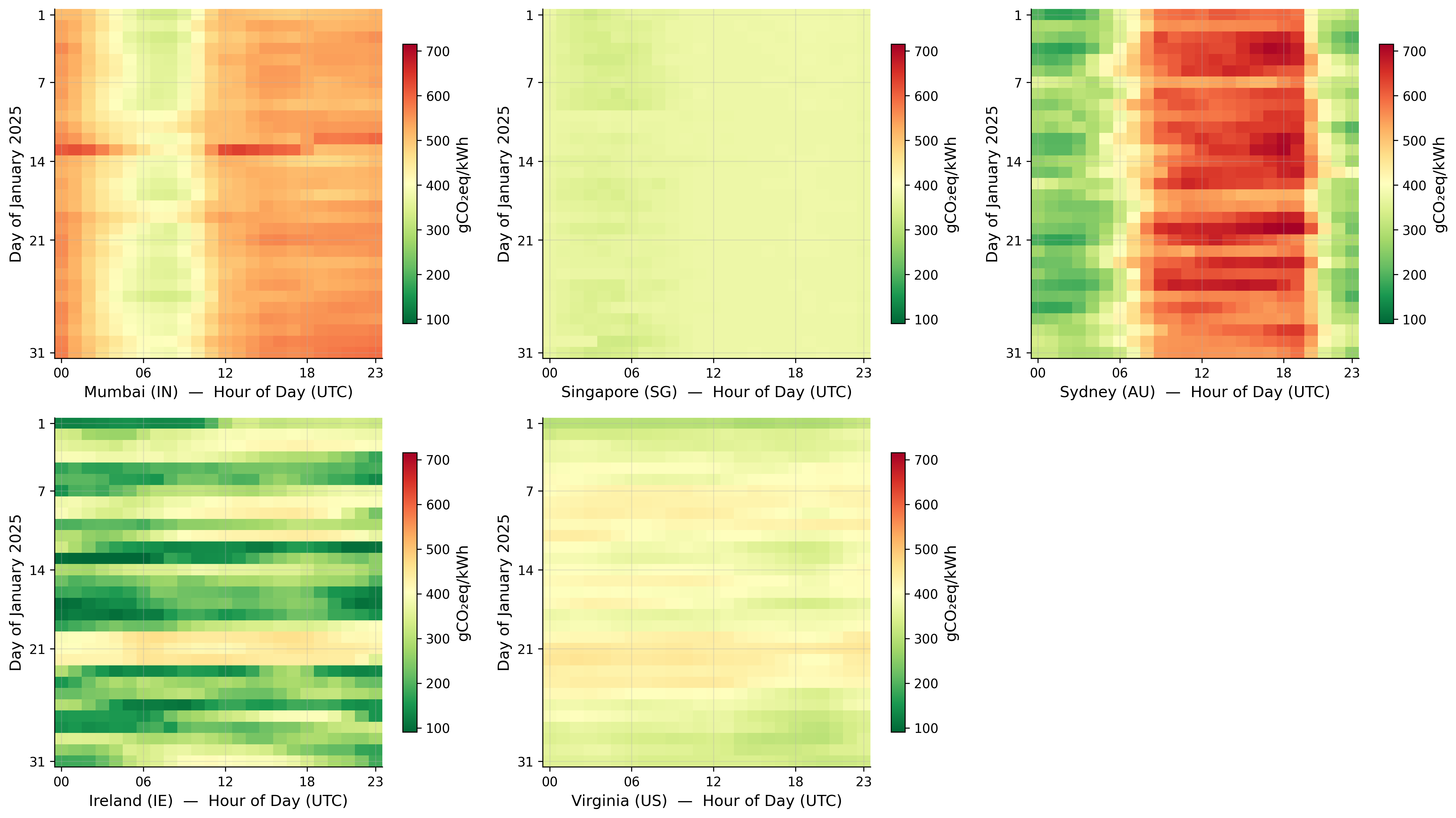}
\caption{Carbon Intensity Heatmap Across Five AWS Regions for January 2025 (Hourly)}
\label{fig:graph1}
\end{figure*}

\begin{figure}[t]
\centering
\includegraphics[height=6.5cm,width=1.0\columnwidth]{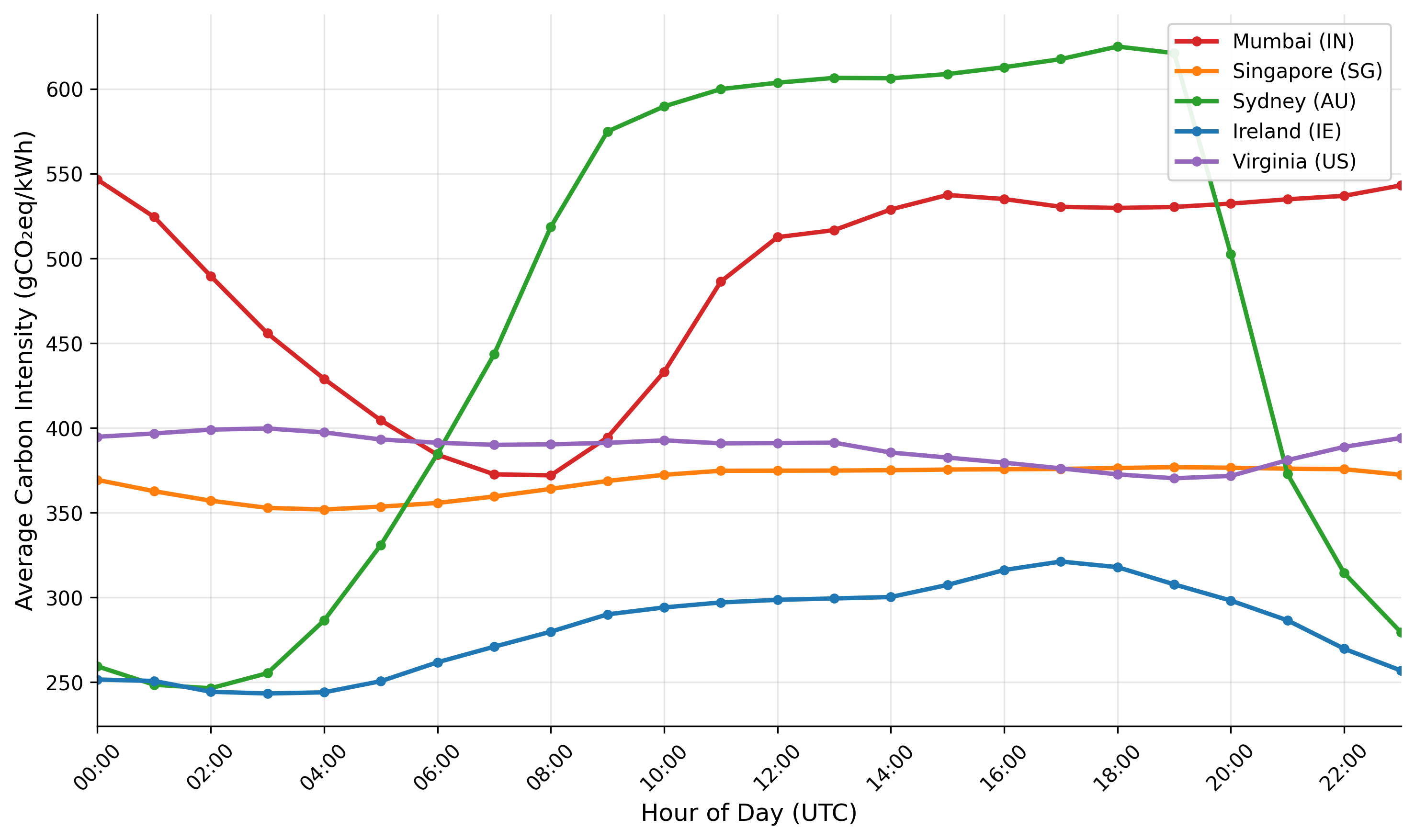}
\caption{Average Carbon Intensity by Hour of Day, Averaged Across January 2025 — A Condensed View of the Daily Pattern.}
\label{fig:graph2}
\end{figure}

Analysis of carbon intensity variation across regions reveals significant regional and temporal differences. Figures \ref{fig:graph1} and \ref{fig:graph2} both illustrate temporal and geographical variability in carbon intensity across AWS regions for January 2025. Where Figure \ref{fig:graph1} presents the full hourly dataset across all 744 hours, while Figure \ref{fig:graph2} condenses it into a daily-averaged profile to make the relative regional ranking easier to compare at a glance. Ireland consistently maintains the lowest carbon intensity due to higher renewable energy contributions, while Mumbai records the highest values. Sydney exhibits noticeable variability, whereas Virginia and Singapore remain relatively stable. These observations confirm that carbon intensity varies significantly across both regions and time, reinforcing the need for carbon-aware routing over static nearest-region selection.

\subsection{Carbon Savings vs SLA Threshold}

As the SLA threshold increases, the carbon savings in the proposed policy C increase gradually. At stringent latency requirements (300 ms), there is no substantial gain in carbon savings because the Mumbai area is the only one that satisfies the SLA requirement. With relaxed SLA constraints, more regions meet the requirement, enabling more carbon savings through carbon-aware routing decisions.

\begin{figure}[t]
\centering
\includegraphics[height=6.5cm,width=1\columnwidth]{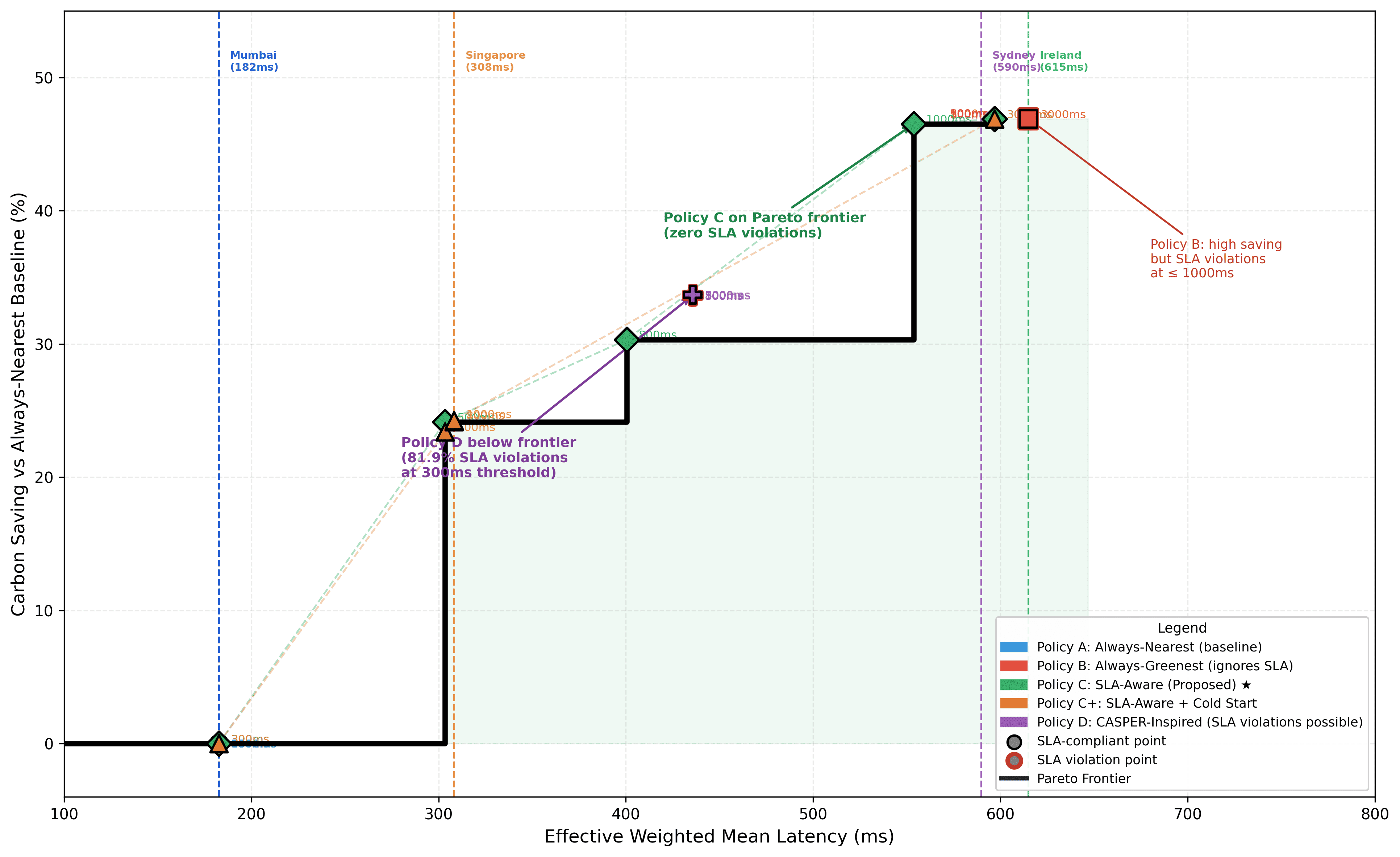}
\caption{Pareto Analysis: Trade-off Between Latency and Carbon Saving Across Five Routing Policies for five primary AWS regions}
\label{fig:pareto}
\end{figure}

Figure \ref{fig:pareto} represents a Pareto analysis of the carbon saving vs effective latency tradeoff across all five policies. A point is Pareto-optimal if no alternative achieves higher carbon savings at equal or lower latency without SLA violations. Policy C occupies positions on the Pareto frontier at two distinct operating points as (308 ms, 30.3\%) and (615 ms, 46.5\%). It demonstrates that no other evaluated policy achieves a superior carbon-latency tradeoff within SLA bounds. Policy D, despite achieving 33.7\% carbon savings, lies strictly below the frontier. Policy C achieves comparable or higher savings at both lower and higher latency points while guaranteeing zero SLA violations.

For example, at an SLA level of 1000 ms, Policy C achieves a reduction of 46.5\% in carbon, while at 3000 ms, Policy C's reduction reaches 46.8\%, approaching the theoretical maximum. Across all SLA thresholds, Policy C maintains 0\% SLA violations, reflecting that carbon savings can be achieved without compromising user experience.

\begin{figure}[t]
\centering
\includegraphics[height=6.6cm,width=1\columnwidth]{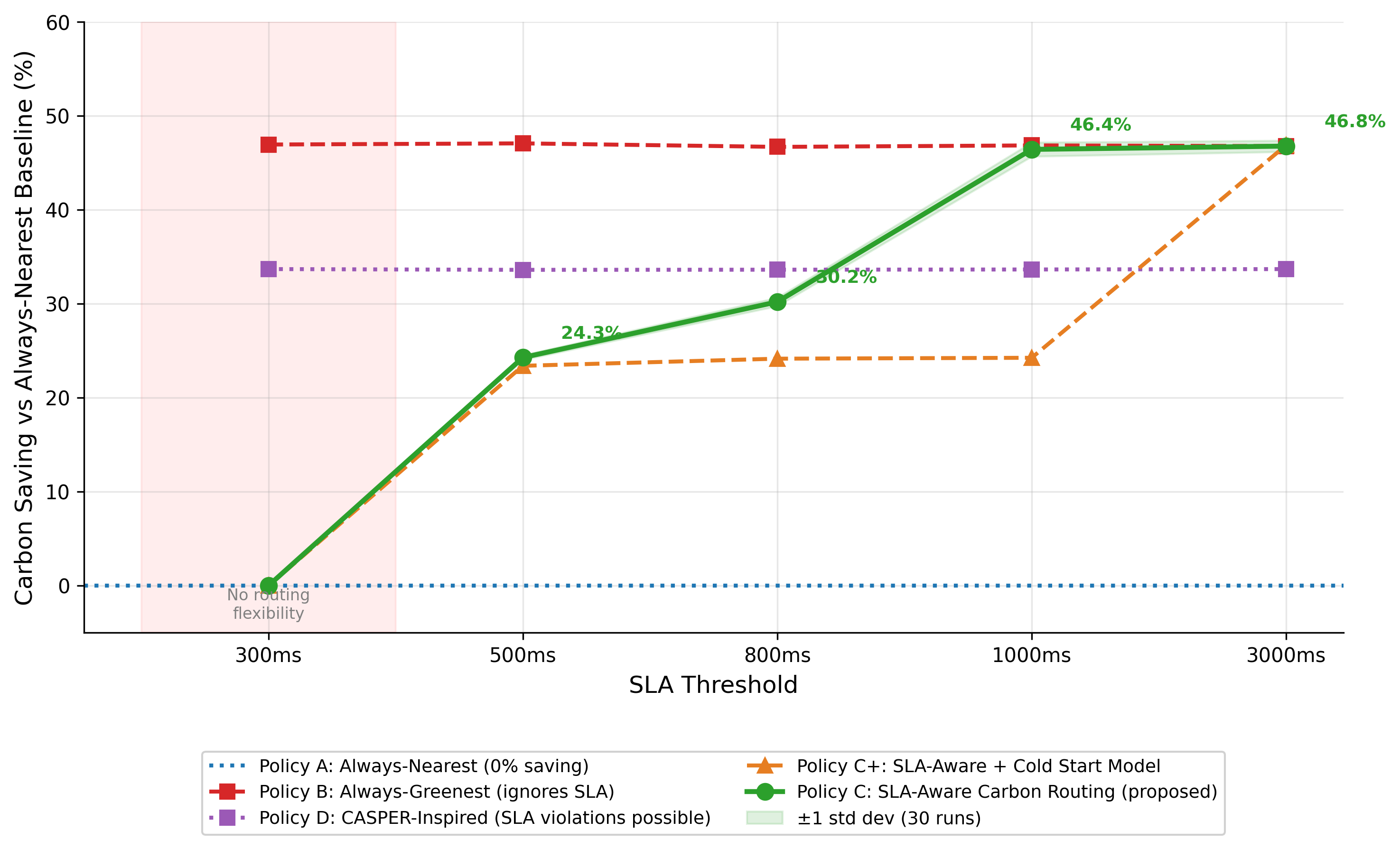}
\caption{\textbf{Carbon Savings Trend Across SLA Thresholds, Illustrating the Divergence Between Hard-Constraint (Policy C) and Soft-Scoring (Policy D) Strategies as Latency Tolerance Increases}}
\label{fig:graph3}
\end{figure}

To ensure robustness with realistic deployment considerations, Policy C+ uses empirical measurements of cold start penalties as inputs to the latency model. In Figure \ref{fig:graph3}, it can be seen that Policy C+ is the same as Policy C when the latency threshold is either 300 ms or 3000 ms. For other values of the threshold, cold start penalties limit routing options. For instance, with an SLA of 1000 ms, Ireland experiences latency greater than the SLA with a cold start, thus making 81.3\% of requests go to Singapore. Consequently, Policy C+ achieves 24.3\% carbon savings at 1000 ms compared to 46.5\% for Policy C, while maintaining zero SLA violations across all thresholds. These results represent a conservative lower bound, as the fixed $\alpha$ = 0.8 assumption intentionally overestimates cold start penalty to guarantee zero SLA violations even during worst-case region transitions. In production deployments where target regions receive sustained traffic and remain warm, actual C+ savings would exceed those reported here.

Figure \ref{fig:graph3} represents carbon footprint savings from each policy with respect to the baseline policy (Policy A) that always has the nearest constraint applied. In case of policy C, savings increase with the reduction of SLA constraints, whereas policy C+ denotes the impact of cold start penalties at intermediate thresholds. Policy D yields constant savings of about 33.7\%; however, it incurs significant SLA violations (81.9\% at 300 ms, 24.3\% at 500 ms, and 21.9\% at 800 ms), indicating that weighted scoring without strict latency constraints is unsuitable for latency-sensitive applications. In contrast to Policy D, which violates SLA constraints under stricter thresholds, Policy C ensures zero SLA violations while achieving higher carbon savings at relaxed SLA levels. 

\begin{table*}[t]
\centering
\caption{Carbon Savings and SLA Compliance Across Five Routing Policies (Mean $\pm$ Std over 30 runs, $n=$ 30,000 routing decisions per SLA threshold per policy)}
\label{tab:carbon_savings_extended}
\footnotesize
\setlength{\tabcolsep}{3pt}
\renewcommand{\arraystretch}{1}

\begin{tabular}{|c|c|c|c|c|c|c|c|c|c|c|}
\hline
\textbf{SLA} & \textbf{Request} & \textbf{Policy A} & \textbf{Policy C} & \textbf{Saving C} & \textbf{Policy C+} & \textbf{Saving C+} & \textbf{Policy D} & \textbf{Saving D} & \textbf{D Violation} & \textbf{p-value} \\
\textbf{(ms)} & \textbf{Type} & (gCO$_2$/kWh) & (gCO$_2$/kWh) & (\%) & (gCO$_2$/kWh) & (\%) & (gCO$_2$/kWh) & (\%) & (\%) & (C vs A) \\
\hline
300  & Real-Time   & 485.9 & 485.9 & $0.0 \pm 0.00$   & 485.9 & $0.0$  & 322.7 & 33.7 & 81.92 & 0.5030 (N/S)\\
\hline
500  & Interactive & 486.3 & 368.2 & $24.1 \pm 0.28$  & 373.6 & $23.4$ & 323.0 & 33.6 & 24.33 & $<$0.0001 \\
\hline
800  & Moderate    & 485.9 & 339.1 & $30.3 \pm 0.35$  & 369.8 & $24.2$ & 322.4 & 33.6 & 21.87 & $<$0.0001 \\
\hline
1000 & Relaxed     & 485.9 & 260.2 & $46.5 \pm 0.41$  & 368.8 & $24.3$ & 322.7 & 33.7 & 0.00  & $<$0.0001 \\
\hline
3000 & Batch       & 485.5 & 258.3 & $46.8 \pm 0.56$  & 258.3 & $46.8$ & 321.9 & 33.7 & 0.00  & $<$0.0001 \\
\hline
\end{tabular}
\vspace{2pt}
\end{table*}

Table \ref{tab:carbon_savings_extended} presents carbon savings and SLA compliance across all routing policies relative to the always-nearest baseline (Policy A). The results show that Policy C achieves substantial carbon reductions while maintaining zero SLA violations across all thresholds, whereas Policy D attains moderate savings at the cost of significant SLA violations under stricter latency constraints. Policy C+ reflects the impact of cold start penalties, particularly at intermediate SLA levels. Overall, the table highlights the trade-off between carbon efficiency and SLA guarantees. Standard deviation is reported for Policy C as the primary evaluation metric, whereas Policy C+ represents a deterministic worst-case bound under cold start assumptions ($\alpha = 0.8$) and thus exhibits negligible variance by construction.

Stricter SLA requirements limit the flexibility of routing, necessitating a selection of geographically closer but more carbon-intensive regions. With relaxed SLA requirements, less carbon-intensive regions are now feasible, enabling greater emission reductions.

\subsection{Statistical Validation}

Consistency across multiple runs is evaluated, and the results show low variance in carbon savings, indicating high stability. The Mann–Whitney U test validates the significant improvement for all thresholds greater than 300 ms ($p < 0.0001$), computed over $n=30$ independent runs per policy per SLA threshold (30,000 routed requests per condition. For 300 ms, no significant difference is observed ($p = 0.5030$), since there is only one region that meets the SLA requirement.

\begin{figure}[t]
\centering
\includegraphics[height=5cm, width=1\columnwidth]{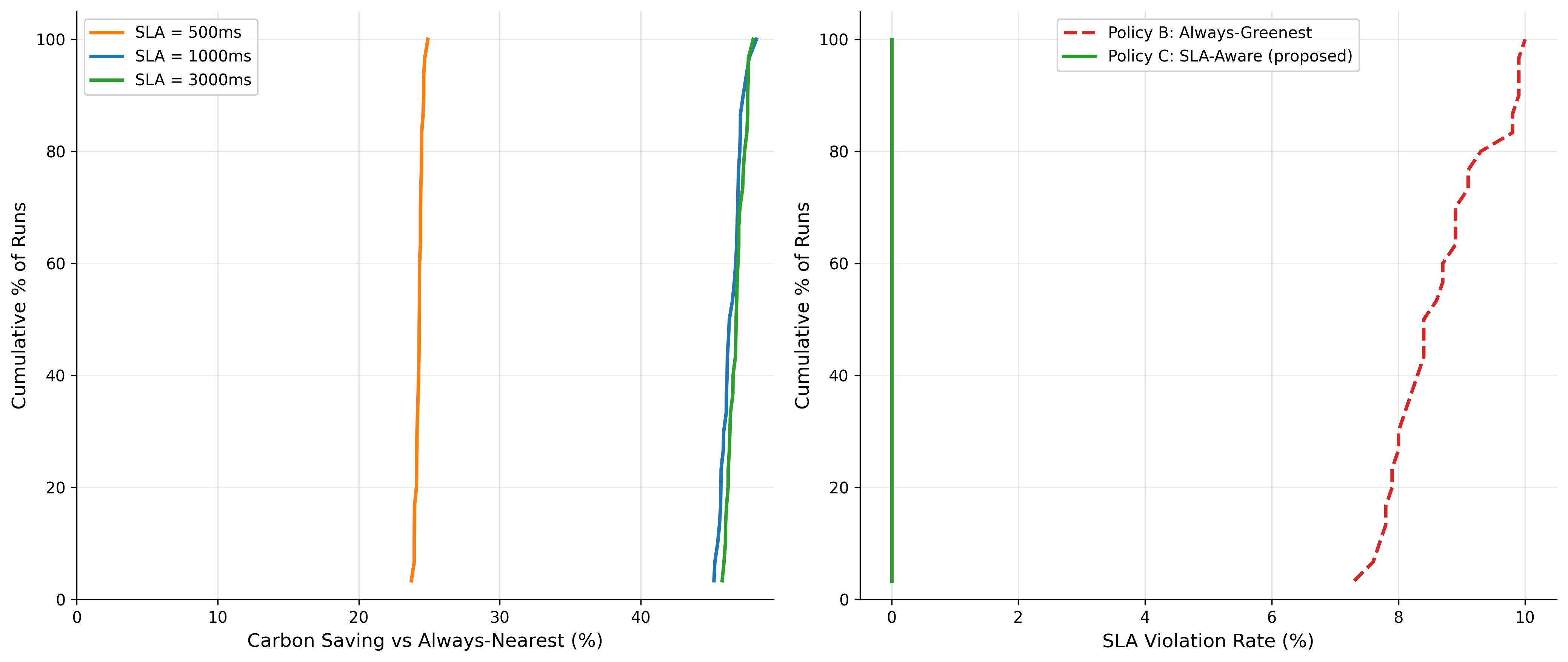}
\caption{Cumulative Distribution Functions of Carbon Saving and SLA Violation Rate}
\label{fig:graph4}
\end{figure}

Figure \ref{fig:graph4} displays the CDF of carbon savings and SLA violation. Policy C shows highly consistent savings with zero SLA violations across all runs, while Policy B exhibits non-zero violations.

\subsection{Routing Behavior Across Regions}

Routing decisions depend on the SLA threshold. Under a strict SLA, routing will be done for regions close to the user. As the SLA constraints relax, routing will take place from lower-carbon-emitting regions like Singapore and Ireland, reflecting adaptive behavior.

\begin{figure}[t]
\centering
\includegraphics[width=1\columnwidth]{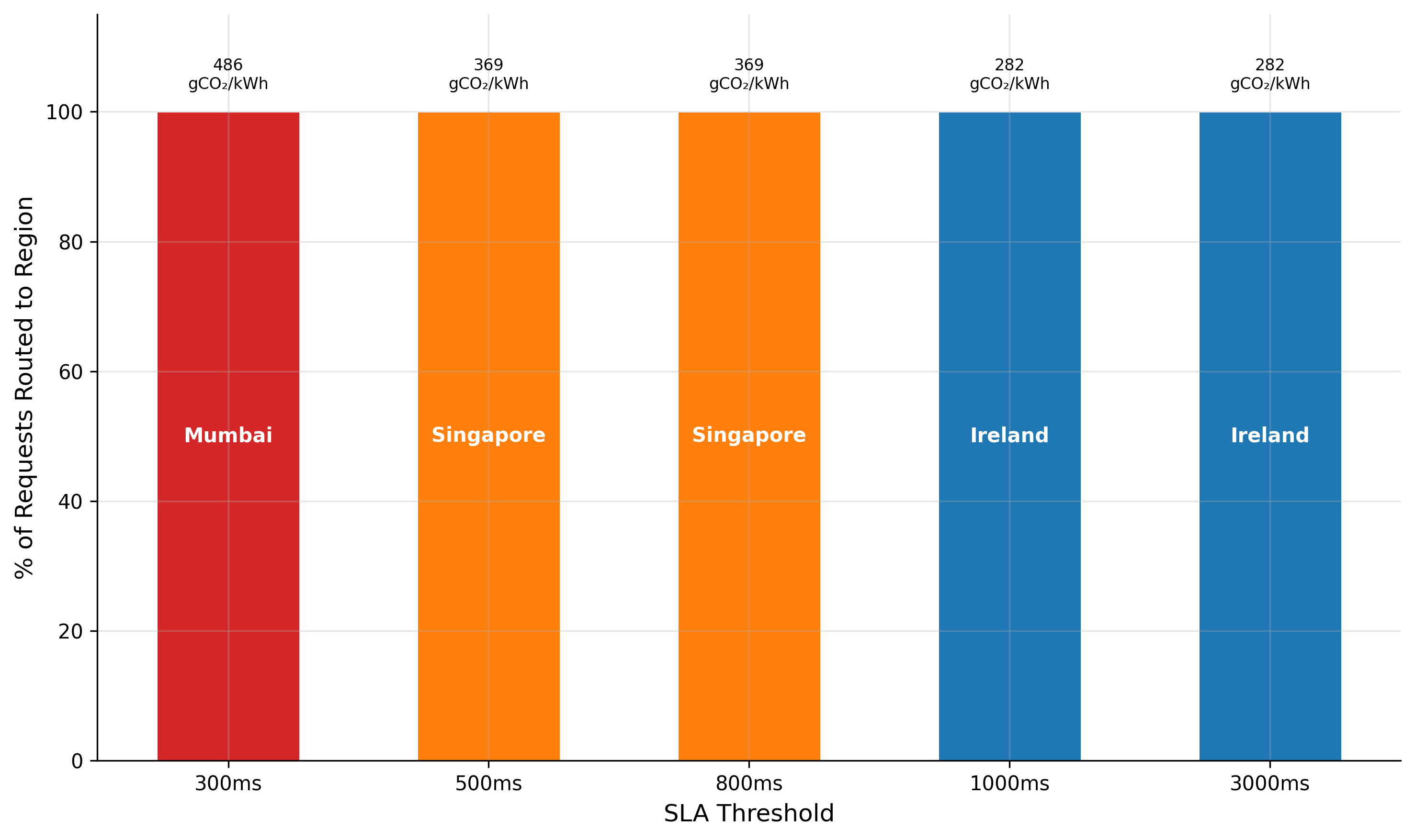}
\caption{Routing Distribution Across AWS Regions Under Policy C (SLA-Aware Routing)}
\label{fig:graph5}
\end{figure}

Figure \ref{fig:graph5} depicts the region-wise distribution of requests. When the delay is 300 ms, routing will be done only to Mumbai. For a less strict SLA, routing will go to Singapore and Ireland, reflecting dynamic adaptation to latency and carbon conditions.

\subsection{Mixed Workload Evaluation}
To evaluate real-world robustness, mixed workloads with heterogeneous SLA requirements are simulated. Under the scenario, the proposed Policy C mitigates carbon emissions by an average of 27.4\% ± 0.27\% while ensuring zero SLA violations. It indicates a weighted aggregation of savings across different SLA classes and effectiveness under realistic traffic conditions.

\begin{figure}[t]
\centering
\includegraphics[height=9cm,width=1\columnwidth]{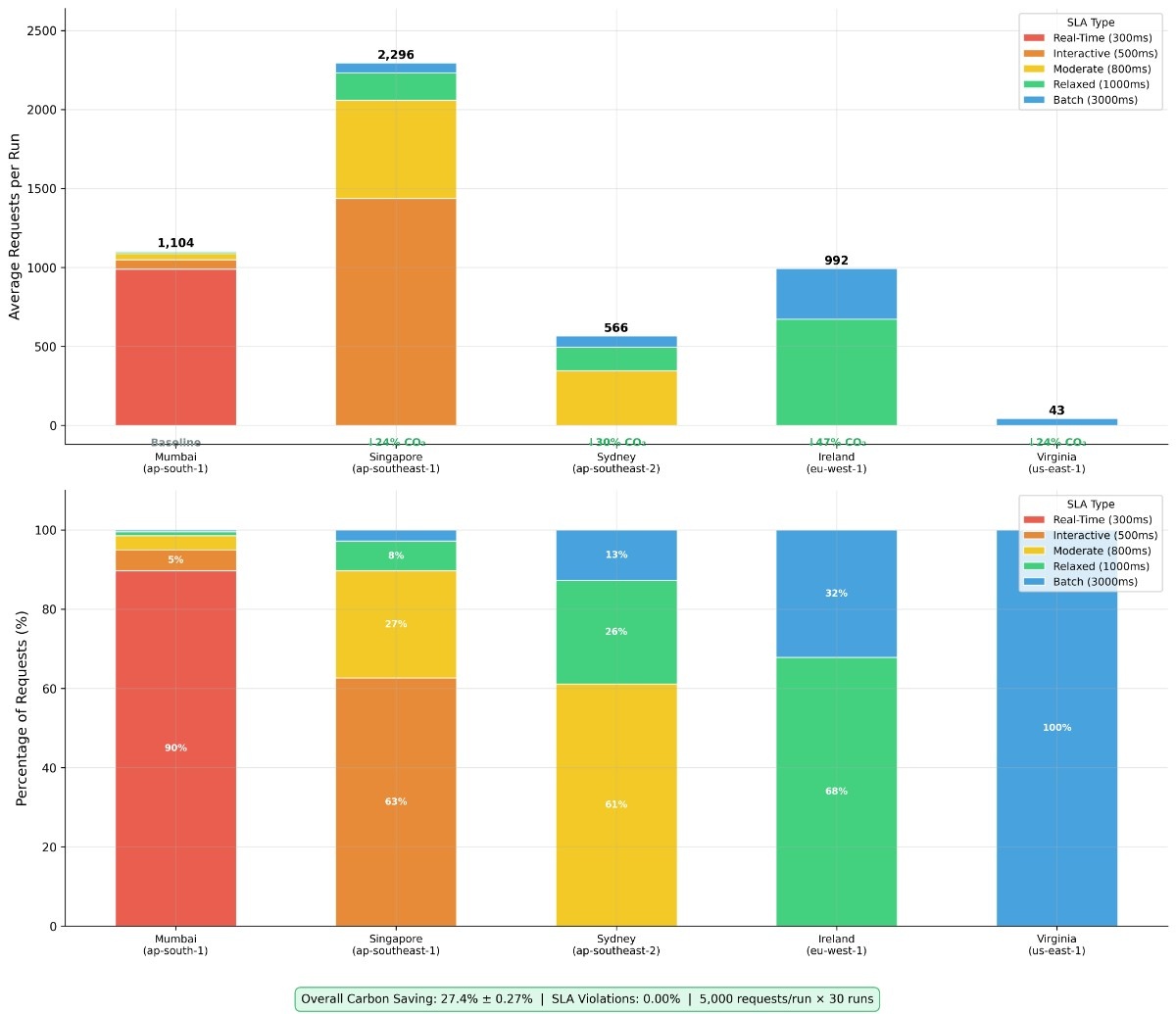}
\caption{Request Distribution per AWS Region Under Mixed Traffic Workload, Stacked by SLA Type}
\label{fig:graph7}
\end{figure}
As illustrated in Figure \ref{fig:graph7}, real-time requests are directed to Mumbai, interactive and moderate loads to Singapore, and relaxed and batch requests primarily to Ireland, yielding 27.4\% ± 0.27\% carbon reduction with zero SLA violations across 150,000 requests.

\begin{table}[t]
\centering
\caption{Mixed Workload Routing Distribution and Carbon Savings}
\label{tab:mixed}
\resizebox{\columnwidth}{!}{%
\small
\begin{tabular}{|l|c|c|l|c|}
\hline
\textbf{SLA Type} & \textbf{Traffic} & \textbf{Reqs} & 
\textbf{Primary Region} & \textbf{Saving} \\ \hline

Real-Time (300ms) & 20\% & 990 & Mumbai (100\%) & 0.0\% \\ \hline
Interactive (500ms) & 30\% & 1,497 & Singapore (96.1\%) & 24.3\% \\ \hline
Moderate (800ms) & 20\% & 1,006 & \shortstack[l]{Singapore (61.8\%)\\+ Sydney (34.4\%)} & 24.3--30.2\% \\ \hline
Relaxed (1000ms) & 20\% & 1,005 & Ireland (67.0\%) & 46.8\% \\ \hline
Batch (3000ms) & 10\% & 503 & Ireland (63.5\%) & 46.8\% \\ \hline
\textbf{Overall} & \textbf{100\%} & \textbf{5,000} & \textbf{Dynamic} & \textbf{27.4\%$\pm$0.27\%} \\ \hline
\end{tabular}%
}
\end{table}

Table \ref{tab:mixed} shows the routing distribution and carbon savings across mixed workload conditions. It illustrates how the proposed approach adapts to heterogeneous SLA requirements while maintaining consistent performance.

\subsection{Routing Overhead}
Finally, the computational overhead introduced by the routing decision process is evaluated. The routing decision leverages a lightweight, in-memory evaluation over precomputed latency and carbon profiles, resulting in negligible overhead. The maximum decision-observed time was just 0.036 ms (P99), accounting for less than 0.02\% of the entire latency of each request. It confirms that the proposed approach is lightweight and practical for real-world deployment.

\subsection{Comparative Analysis Against CASPER Baseline}

The proposed method is benchmarked against Policy D, a weighted-scoring ablation structurally inspired by spatial-shifting approaches such as CASPER, which optimizes carbon emissions without applying hard SLA constraints. The comparison is intended to isolate the effect of constraint enforcement rather than to serve as a faithful reimplementation of CASPER, which incorporates SLO and latency constraints as a first-class part of its own optimization. As shown in Table \ref{tab:carbon_savings_extended}, Policy D achieves a consistent carbon reduction of 33.7\% across all SLA thresholds but incurs SLA violations under strict latency constraints. Policy C, in contrast, achieves zero SLA violations at all thresholds.

One limitation of soft optimization techniques is that routing decisions may violate latency requirements without explicit constraints, whereas Policy C provides hard SLA guarantees at all levels. Policy C outperforms Policy D in carbon reduction while service reliability is maintained for looser SLA demands (1000 ms and higher) which shows the strength of filter-then-minimize strategy for latency-sensitive applications against weighted scoring methods.

\subsection{Scalability Analysis}

To evaluate scalability, the deployment is extended from 5 to 12 AWS regions spanning 6 continents under mixed workload conditions. The AWS regions augmentations improves mean carbon savings from 27.4\% ± 0.27\% to 47.5\% ± 0.11\% (a 20.1 percentage point gain) while maintaining zero SLA violations. The observed improvement results from enhanced routing flexibility, which allows access to geographically distant, low-carbon regions under relaxed latency constraints. Conversely, performance under strict service-level agreement (SLA) thresholds remains unchanged due to inherent latency limitations, indicating that scalability does not compromise strict SLA guarantees. Stockholm [eu-north-1], with an approximate carbon intensity of 15 (gCO$_2$/kWh), is a significant contributor to these gains. The analysis expands the routing pool rather than evaluating algorithmic scalability; the routing decision remains $O(|R|)$ and completes in under 0.036 ms regardless of the number of regions.

\begin{figure*}[t]
\centering
\includegraphics[width=1.0\textwidth]{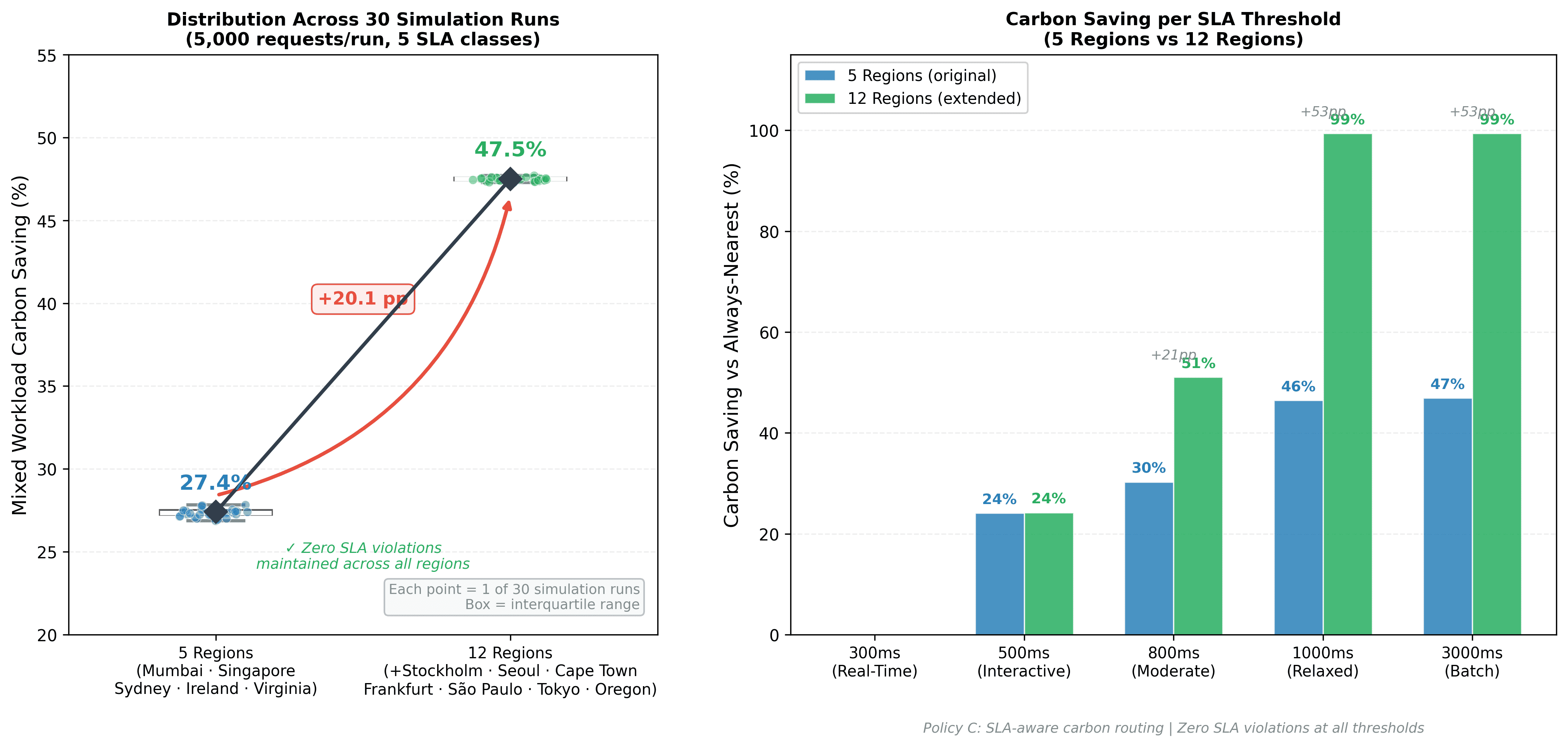}
\caption{Scalability Analysis: Mixed Workload Carbon Saving as 
AWS Regions Scale from 5 to 12 Across 6 Continents.}
\label{fig:scale}
\end{figure*}

Figure \ref{fig:scale} shows the distribution of mixed workload carbon savings over 30 runs. Increasing the number of AWS regions from 5 to 12, augmenting the mean carbon saving from 27.4\% ± 0.27\% to 47.5\% ± 0.11\%, reflecting the consistent gains with low variance and zero SLA violations. At strict SLAs (300-500 ms), the improvements are small due to the latency constraints, but at relaxed thresholds (800-3000 ms), the improvements grow significantly as more low-carbon regions, like Stockholm, become eligible. These results confirm that the proposed framework scales efficiently with infrastructure expansion while ensuring SLA guarantees for all latency thresholds.

\section{Conclusion and Future Directions}

The proposed work focuses on an SLA-aware carbon-routing framework for serverless applications deployed across geographically distributed cloud regions. Unlike conventional latency-aware routing, the proposed model treats SLA constraints as a hard feasibility boundary rather than a weighted objective function. The proposed model selects the least carbon-intensive region within the feasible set while satisfying per-request SLA constraints. The framework is evaluated across 5 AWS regions and further extended to 12 regions using real-world carbon and latency data, spanning 6 continents. It achieves carbon savings of up to 46.8\% with an average of 27.4\% under mixed workloads, with zero SLA violations and negligible overhead. Additionally, it achieves up to 47.5\% carbon savings with expanded routing flexibility. A cold start-aware variant (Policy C+) maintains zero SLA violations while maintaining substantial carbon savings. These results show that significant carbon savings can be achieved without modifying application or infrastructure logic. However, the current evaluation relies on static carbon datasets and fixed workload intensity. It provides per-decision feasibility guarantees but does not establish formal bounds on long-term carbon optimality.

Future work includes introducing real-time carbon and latency measurements, performance analysis under dynamic workloads, cost-carbon analysis, improvement in cold-start management, deployment of the proposed framework to multi-cloud and edge computing environments, and validation of the approach across geographically diverse client locations.

\bibliographystyle{ACM-Reference-Format}
\bibliography{sample-base}

\end{document}